\documentclass[letterpaper, 10 pt, conference]{ieeeconf}  %

\IEEEoverridecommandlockouts                              %

\usepackage{amsmath}
\usepackage{graphicx}
\usepackage{subfig}
\usepackage{tabularx}
\usepackage{booktabs} %
\usepackage{cite}
\usepackage[pagebackref=true,breaklinks=true,letterpaper=true,colorlinks,bookmarks=false]{hyperref}

\usepackage[margin=4pt,font=footnotesize,labelfont=bf,labelsep=endash,tableposition=top]{caption}

\DeclareMathAlphabet{\pazocal}{OMS}{zplm}{m}{n}

\newcommand\T{\rule{0pt}{2.2ex}}       %
\newcommand\B{\rule[-0.8ex]{0pt}{0pt}} %
\title{\LARGE \bf
Real-Time Joint Semantic Segmentation and Depth Estimation Using Asymmetric Annotations
}

\author{Vladimir Nekrasov$^{1}$, Thanuja Dharmasiri$^{2}$, Andrew Spek$^{2}$, Tom Drummond$^{2}$, Chunhua Shen$^{1}$ and Ian Reid$^{1}$%
\thanks{$^{1}$School of Computer Science, the University of Adelaide, Australia
        {\tt\small \{firstname.lastname\}@adelaide.edu.au}}%
\thanks{$^{2}$Monash University, Australia \newline
        {\tt\small \{firstname.lastname\}@monash.edu}}%
}

\begin{document}

\newcolumntype{Y}{>{\centering\arraybackslash}p}
\def\etal{\emph{et al.}}

\maketitle
\thispagestyle{empty}
\pagestyle{empty}

\begin{abstract}
	Deployment of deep learning models in robotics as sensory information extractors can be a daunting task to handle, even using generic GPU cards. Here, we address three of its most prominent hurdles, namely, i) the adaptation of a single model to perform multiple tasks at once (in this work, we consider depth estimation and semantic segmentation crucial for acquiring geometric and semantic understanding of the scene), while ii) doing it in real-time, and iii) using asymmetric datasets with uneven numbers of annotations per each modality. To overcome the first two issues, we adapt a recently proposed real-time semantic segmentation network, making changes to further reduce the number of floating point operations. To approach the third issue, we embrace a simple solution based on hard knowledge distillation under the assumption of having access to a powerful `teacher' network. We showcase how our system can be easily extended to handle more tasks, and more datasets, all at once, performing depth estimation and segmentation both indoors and outdoors with a single model. Quantitatively, we achieve results equivalent to (or better than) current state-of-the-art approaches with one forward pass costing just 13ms and 6.5~GFLOPs on 640$\times$480 inputs. This efficiency allows us to directly incorporate the raw predictions of our network into the SemanticFusion framework~\cite{McCormacHDL17} for dense 3D semantic reconstruction of the scene.%
	\footnote[3]{The models are available here: {\url{https://github.com/drsleep/multi-task-refinenet}}}
\end{abstract}

\section{INTRODUCTION}

As the number of tasks on which deep learning shows impressive results continues to grow in range and diversity, the number of models that achieve such results keeps analogously increasing, making it harder for practitioners to deploy a complex system that needs to perform multiple tasks at once. For some closely related tasks, such a deployment does not present a significant obstacle, as besides structural similarity, those tasks tend to share the same datasets, as, for example, the case of image classification, object detection, and semantic segmentation. On the other hand, tasks like segmentation and depth estimation %
rarely (fully) share the same dataset;
for example, the NYUD dataset~\cite{SilbermanHKF12,GuptaAM13} comprises a large set of annotations for depth estimation, %
but only a small labelled set of segmentations. One can readily approach this problem by simply updating the parameters of each task only if there exist ground truth annotations for that task. Unfortunately, this often leads to suboptimal results due to imbalanced and biased gradient updates. We note that while it is not clear how to handle such a scenario in the most general case, in this paper we assume that we have access to a large and powerful model, that can make an informative prediction to acquire missing labels. For each single task considered separately, this assumption is often-times valid, and we make use of it to predict missing segmentation masks.

Another issue that arises is the imperative in the context of robotics and autonomous systems for extraction of sensory information in real time.
While there has been a multitude of successful approaches to speed up individual tasks~\cite{RenHG015,IandolaMAHDK16,ZhaoQSSJ17}, there is barely any prior work on performing multiple tasks concurrently in real-time. Here we show how to perform two tasks, depth estimation and semantic segmentation, in real-time with very few architectural changes and without any complicated pipelines.

Our choice of tasks is motivated by an observation that, for all sorts of robotic applications it is important for a robot (an agent) to know the semantics of its surroundings and to perceive the distances to the surfaces in the scene. The proposed methodology is simple and achieves competitive results in comparison to large models. Furthermore, we believe that there is nothing that prohibits practitioners and researchers to adapt our method for more tasks, which, in turn, would lead to better exploitation of deep learning models in real-world applications. To confirm this claim, we conduct additional experiments, predicting besides depth and segmentation, surface normals. Moreover, we successfully train a single model able to perform depth estimation and semantic segmentation, together in both indoor and outdoor settings. In yet another case study, we demonstrate that raw outputs of our joint network (segmentation and depth) can be directly used inside the SemanticFusion framework~\cite{McCormacHDL17} to estimate dense semantic $3$D reconstruction of the scene in real-time.

To conclude our introduction, we re-emphasise that our results demonstrate that there is no need to uncritically deploy multiple expensive models, when the same performance can be achieved with one small network - a case of one being better than two!

\begin{figure*}
	\centering
	\includegraphics[width=\textwidth]{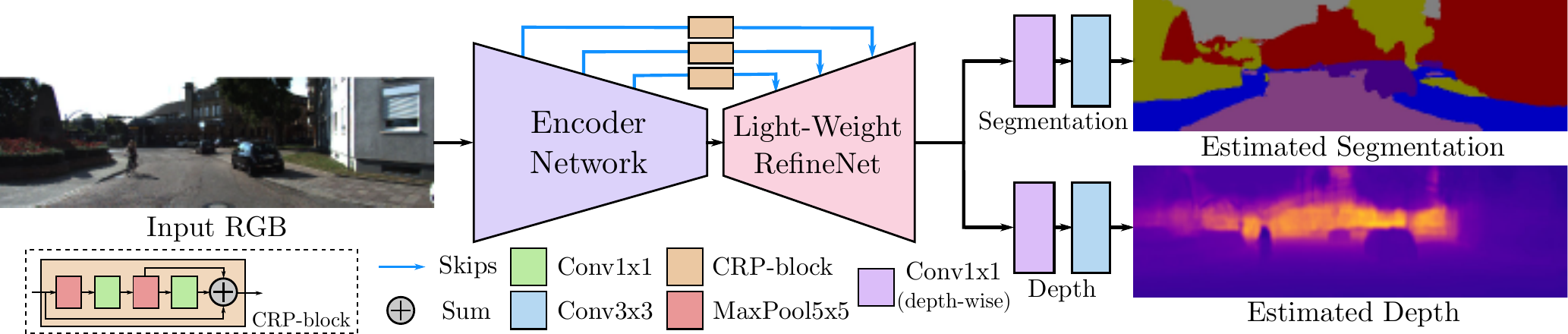}
	\caption{General network structure for joint semantic segmentation and depth estimation. Each task has only $2$ specific parametric layers, while everything else is shared}
	\label{fig:arch}
	\vskip -0.2in
\end{figure*}

\section{Related Work}
\label{sec:rel_work}

Our work is closely related to several topics. Among them are multi-task learning, semantic segmentation, depth estimation, and knowledge distillation.

According to the classical {\em multi-task learning} paradigm, forcing a single model to perform several related tasks simultaneously can improve generalisation via imposing an inductive bias on the learned representations~\cite{Caruana93,Baxter00}. Such an approach assumes that all the tasks use a shared representation before learning task-specific parameters. Multiple works in computer vision have been following this strategy; in particular, Eigen \& Fergus~\cite{EigenF15} trained a single architecture (but with different copies) to predict depth, surface normals and semantic segmentation, Kokkinos~\cite{Kokkinos17} proposed a universal network to tackle $7$ different vision tasks, Dvornik ~\etal\cite{DvornikSMS17} found it beneficial to do joint semantic segmentation and object detection, while Kendall~\etal\cite{KendallGC17} learned optimal weights to perform instance segmentation, semantic segmentation and depth estimation all at once. Chen~\etal~\cite{ChenYML18} built a single network with the ResNet-50~\cite{HeZRS16} backbone performing joint semantic segmentation, depth estimation and object detection. To alleviate the problem of imbalanced annotations, Kokkinos~\cite{Kokkinos17} chose to accumulate the gradients for each task until a certain number of examples per task is seen, while Dvornik~\etal \cite{DvornikSMS17} simply resorted to keeping the branch with no ground truth available intact until at least one example of that modality is seen.

We note that none of these methods makes any use of already existing models for each separate task, and none of them, with the exception of BlitzNet~\cite{DvornikSMS17}, achieves real-time performance. In contrast, we show how to exploit large pre-trained models to acquire better results, and how to do inference in real-time.

{\em Semantic segmentation} is a task of per-pixel label classification, and most approaches in recent years have been centered around the idea of adapting image classification networks into fully convolutional ones able to operate on inputs of different sizes~\cite{LongSD15, WuSH16e, ChenPSA17}. Real-time usage of such networks with decent performance is a non-trivial problem, and few approaches are currently available~\cite{ZhaoQSSJ17, PaszkeCKC16, LiLLLT17, nekrasovlight}. We have chosen recently proposed Light-Weight RefineNet~\cite{nekrasovlight} on top of MobileNet-v2~\cite{abs-1801-04381} as our baseline architecture as it exhibits solid performance on the standard benchmark dataset, PASCAL VOC~\cite{EveringhamGWWZ10} in real-time, while having fewer than $4$M parameters.

{\em Depth estimation} is another per-pixel task, the goal of which is to determine how far each pixel is from the observer. Traditionally, image based depth reconstruction was performed using SLAM based approaches~\cite{Newcombe2011,engel2014lsd,klein2007parallel}. However, recent machine learning approaches have achieved impressive results, where a CNN has been successfully employed to predict a depth map from a single RGB image using supervised learning~\cite{EigenF15,EigenNIPS,Laina2016,Liu}, unsupervised learning \cite{garg2016unsupervised,godard2017unsupervised} and semi-supervised learning \cite{kuznietsov2017semi}. Predicting multiple quantities including depths from a single image was first tackled by Eigen \& Fergus~\cite{EigenF15}. Dharmasiri \etal\cite{dharmasiri2017joint} demonstrated that predicting related structural information in the form of depths, surface normals and surface curvature results in improved performances of all three tasks compared to utilising three separate networks. Most recently, Qi~\etal~\cite{qi2018geonet} found it beneficial to directly encode a geometrical structure as part of the network architecture in order to perform depth estimation and surface normals estimation simultaneously. Our approach is fundamentally different to these previous works in two ways. Firstly, our network exhibits real-time performance on each individual task. Secondly, we demonstrate how to effectively incorporate asymmetric and uneven ground truth annotations into the training regime. Furthermore, it should be noted that despite using a smaller model running in real-time, we still quantitatively outperform these approaches.

Finally, we briefly touch upon the {\em knowledge distillation} approach~\cite{BucilaCN06,HintonVD15,BaC14,RomeroBKCGB14} that is based on the idea of having a large pre-trained teacher (expert) network (or an ensemble of networks), and using its logits, or predictions directly, as a guiding signal for a small network along with original labels. Several previous works relied on knowledge distillation to either acquire missing data~\cite{zamir2018taskonomy}, or as a regulariser term~\cite{hoffman2016learning,li2017learning}. While those are relevant to our work, we differ along several axes: most notably, Zamir~\etal~\cite{zamir2018taskonomy} require separate network copies for different tasks, while Hoffman~\etal~\cite{hoffman2016learning} and Li \& Hoiem~\cite{li2017learning} only consider a single task learning (object detection and image classification, respectively).

\section{Methodology}
While we primarily discuss the case with only two tasks present, the same machinery applies for more tasks, as demonstrated in Sect.~\ref{subsec:ext1}.
\label{sec:method}

\subsection{Backbone Network}
As mentioned in the previous section, we employ the recently proposed Light-Weight RefineNet architecture~\cite{nekrasovlight} built on top of the MobileNet-v2 classification network~\cite{abs-1801-04381}. This architecture extends the classifier by appending several simple contextual blocks, called Chained Residual Pooling (CRP)~\cite{LinMSR17}, consisting of a series of $5\times5$ max-pooling and $1\times1$ convolutions (Fig.~\ref{fig:arch}).

Even though the original structure already achieves real-time performance and has a small number of parameters, for the joint task of depth estimation and semantic segmentation (of $40$ classes) it requires more than $14$~GFLOPs on inputs of size $640\times480$, which may hinder it from the direct deployment on mobile platforms with few resources available. We found that the last CRP block is responsible for more than half of the FLOPs as it deals with the high-resolution feature maps ($1/4$ from the original resolution). Thus, to decrease its influence, we replace $1\times1$ convolution in the last CRP block with its depthwise equivalent (i.e., into a grouped convolution with the number of groups being equal to the number of input channels)~\cite{Chollet17}. Doing so reduces the number of operations by more than half, down to just $6.5$ GFLOPs.%

\subsection{Joint Semantic Segmentation and Depth Estimation}
In the general case, it is non-trivial to decide where to branch out the backbone network into separate task-specific paths in order to achieve the optimal performance on all of them simultaneously. For simplicity, we branch out right after the last CRP block, and append two additional convolutional layers (one depthwise $1\times1$ and one plain $3\times3$) for each task (Fig.~\ref{fig:arch}).

If we denote the output of the network before the branching as $\tilde{y}=f_{\theta_{b}}{(I)}$, where $f_{\theta_{b}}$ is the backbone network with a set of parameters $\theta_{b}$, and $I$ is the input RGB-image, then the depth and segmentation predictions can be denoted as $\tilde{y}_{s}=g_{\theta_{s}}{(\tilde{y})}$ and $\tilde{y}_{d}=g_{\theta_{d}}{(\tilde{y})}$, where $g_{\theta_{s}}$ and $g_{\theta_{d}}$ are segmentation and depth estimation branches with the sets of parameters $\theta_{s}$ and $\theta_{d}$, respectively. We use the standard softmax cross-entropy loss for segmentation and the inverse Huber loss for depth estimation~\cite{Laina2016}. Our total loss (Eqn.~(\ref{eqn1})) contains an additional scaling parameter, $\lambda$, which, for simplicity, we set to $0.5$:

\setlength{\arraycolsep}{0.0em}
\begin{align}
\begin{split}
\label{eqn1}
&\pazocal{L}_{total}(I, G_{s}, G_{d}; \theta_{b}, \theta_{s}, \theta_{d}) = (\lambda \cdot \pazocal{L}_{segm}(I, G_{s}; \theta_{b}, \theta_{s})~ + \\
& ~~~~~~~~~~~~~~~ (1 - \lambda) \cdot \pazocal{L}_{depth}(I, G_{d}; \theta_{b}, \theta_{d})),\\
&\pazocal{L}_{segm}(I, G)=\frac{-1}{|I|}\sum_{i\in{I}}log(softmax(\tilde{y}_{s})_{iG_{i}}), \\
\
&\pazocal{L}_{depth}(I, G)= 
\begin{cases}
|\tilde{y}_{d} - G|,~\text{if}~|\tilde{y}_{d} - G| \leq c\\
((\tilde{y}_{d} - G)^{2} + c^{2}) / (2c), ~ \text{otherwise},\\
\end{cases}
\\
&~c\stackrel{\text{def}}{=}0.2 \cdot \max{|\tilde{y}_{d} - G|},\\
\end{split}
\setlength{\arraycolsep}{5pt}
\end{align}
where $G_{s}$ and $G_{d}$ denote ground truth segmentation mask and depth map, correspondingly; $(\cdot)_{ij}$ in the segmentation loss is the probability value of class $j$ at pixel $i$.

\setlength{\tabcolsep}{4pt}
\begin{table*}[htb]
	\begin{center}
	\vskip 0.05in
	\caption{Results on the test set of NYUDv2. The speed of a single forward pass and the number of FLOPs are measured on $640\times480$ inputs. For the reported mIoU the higher the better, whereas for the reported RMSE the lower the better. ($\dagger$) means that both tasks are performed simultaneously using a single model, while ($\ddagger$) denotes that two tasks employ the same architecture but use different copies of weights per task
			\label{table:nyud1}}
			\begin{tabular}{l|c|c|c|c|c|c|c}
				\specialrule{.15em}{0em}{0em} 
				&&\textbf{Sem. Segm.} & \multicolumn{2}{c|}{\textbf{Depth Estimation}} & \multicolumn{3}{|c}{\textbf{General}}\T\B\\
				\specialrule{.1em}{0em}{0em}
				Model & Regime & mIoU,\% & RMSE (lin),m & RMSE (log) & Parameters,M & GFLOPs & speed,ms (mean/std)\T\B\\
				\specialrule{.1em}{0em}{0em}
				$\dagger$\textbf{Ours} & Segm,Depth & $42.02$ & $0.565$ & $0.205$ & \textbf{3.07} & \textbf{6.49} & \textbf{12.8$\pm$0.1}\T\B\\
				\hline
				RefineNet-101~\cite{LinMSR17} & Segm & \textbf{43.6} & $-$ & $-$ & $118$ & $-$ & $60.3\pm0.5$\T\\
				RefineNet-LW-50~\cite{nekrasovlight} & Segm & $41.7$ & $-$ & $-$ & $27$ & $33$ & $19.6\pm0.3$\\
				Context~\cite{LinSRH15} & Segm & $40.6$ & $-$ & $-$ & $-$ & $-$ & $-$\\
				$\dagger$Sem-CRF+~\cite{MousavianPK16} & Segm,Depth & $39.2$ & $0.816$ & $0.314$ & $-$ & $-$ & $-$\\
				$\ddagger$Kendall and Gal~\cite{KendallGal2017Uncertainties}& Segm,Depth & $37.3$ & \textbf{0.506} & $-$ & $-$ & $-$ & $150$\\
				Fast Res.Forests~\cite{ZuoD17} & Segm & $34.3$ & $-$ & $-$ & $-$ & $-$ & $48.4$\\
				$\ddagger$Eigen and Fergus~\cite{EigenF15} & Segm,Depth & $34.1$ & $0.641$ & $0.214$ & $-$ & $-$ & $-$\\
				Laina~\etal~\cite{Laina2016} & Depth & $-$ & $0.573$ & \textbf{0.195} & $63.6$ & $-$ & $55$\\
				$\dagger$Qi~\etal~\cite{qi2018geonet} & Depth,Normals & $-$ & $0.569$ & - & - & $-$ & $870$\B\\
				\specialrule{.15em}{0em}{0em}
			\end{tabular}
	\end{center}
	\vskip -0.2in
\end{table*}
\setlength{\tabcolsep}{1.4pt}

\subsection{Expert Labeling for Asymmetric Annotations}
\label{subsec:exp}
As one would expect, it is impossible to have all the ground truth sensory information available for each single image. Quite naturally, this poses a question of how to deal with a set of images $S=\{I\}$ among which some have an annotation of one modality, but not another. Assuming that one modality is always present for each image, this then divides the set $S$ into two disjoint sets $S_{1}=S_{T_{1}}$ and $S_{2}=S_{T_{1},T_{2}}$ such that $S=S_{1}\cup{S_{2}}$, where $T_{1}$ and $T_{2}$ denote two tasks, respectively, and the set $S_{1}$ consists of images for which there are no annotations of the second task available, while $S_{2}$ comprises images having both sets of annotations.

Plainly speaking, there is nothing that prohibits us from still exploiting equation~(\ref{eqn1}), in which case only the weights of the branch with available labels will be updated. As we show in our experiments, this leads to biased gradients and, consequently, sub-optimal solutions. Instead, emphasising the need of updating both branches simultaneously, we rely on an expert model to provide us with noisy estimates in place of missing annotations.%

More formally, if we denote the expert model on the second task as $E_{T_{2}}$, then its predictions $\tilde{S}_{1}=E_{T_{2}}(S_{1})$ on the set $S_{1}$ can be used as synthetic ground truth data, which we will use to pre-train our joint model before the final fine-tuning on the original set $S_{2}$ with readily available ground truth data for both tasks. Here, we exploit the labels predicted by the expert network instead of logits, as storing a set of large $3$-D floating point tensors requires extensive resources.

Note also that our framework is directly transferable to cases when the set $S$ comprises several datasets. %
In Sect.~\ref{subsec:ext2} we showcase a way of exploiting all of them in the same time using a single copy of the model.

\section{Experimental Results}
\label{sec:result}

In our experiments, we consider two datasets, NYUDv2-40~\cite{SilbermanHKF12,GuptaAM13} and KITTI~\cite{GeigerLSU13,ros:2015}, representing indoor and outdoor settings, respectively, and both being used extensively in the robotics community.

All the training experiments follow the same protocol. In particular, we initialise the classifier part using the weights pre-trained on ImageNet~\cite{DengDSLL009}, and train using mini-batch SGD with momentum with the initial learning rate of $1e$-$3$ and the momentum value of $0.9$. Following the setup of Light-Weight RefineNet~\cite{nekrasovlight}, we keep batch norm statistics frozen. We divide the learning rate by $10$ after pre-training on a large set with synthetic annotations. We train with a random square crop of $350\times350$ augmented with random mirroring.

All our networks are implemented in PyTorch~\cite{paszke2017automatic}. To measure the speed performance, we compute $100$ forward passes and report both the mean and standard deviation values, as done in~\cite{nekrasovlight}. Our workstation has $24$GB RAM, Intel i5-7600 processor and a single GT1080Ti GPU card running CUDA9.0 and CuDNN7.0.

\subsection{NYUDv2}
\label{exp-nyudv2}

NYUDv2 is an indoor dataset with $40$ semantic labels. It contains $1449$ RGB images with both segmentation and depth annotations, of which $795$ comprise the training set and $654$ - validation. The raw dataset contains more than $300,000$ training images with depth annotations. During training we use less than $10\%$ ($25$K images) of this data. As discussed in Sect.~\ref{subsec:exp}, we annotate these images for semantic segmentation using a teacher network (here, we take the pre-trained Light-Weight RefineNet-152~\cite{nekrasovlight} that achieves $44.4\%$ mean iou on the validation set). After acquiring the synthetic annotations, we pre-train the network on the large set, and then fine-tune it on the original small set of $795$ images.

\begin{figure}[htb]
	\centering
	\resizebox{0.49\textwidth}{!}{
		\begin{tabular}{ccccc}
			\subfloat{\includegraphics[width = 0.19\linewidth]{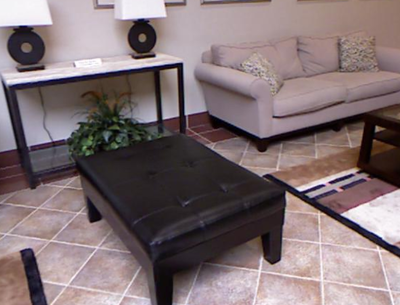}} &
			\subfloat{\includegraphics[width = 0.19\linewidth]{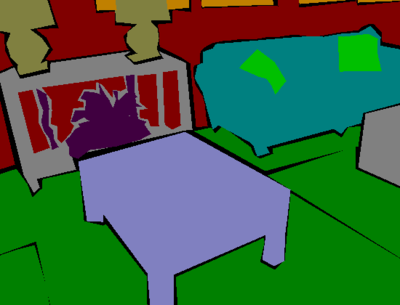}} &
			\subfloat{\includegraphics[width = 0.19\linewidth]{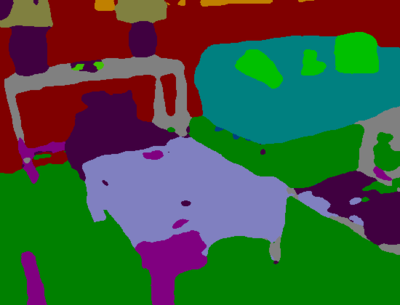}} &
			\subfloat{\includegraphics[width = 0.19\linewidth]{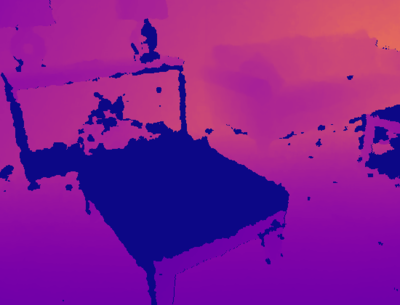}} &
			\subfloat{\includegraphics[width = 0.19\linewidth]{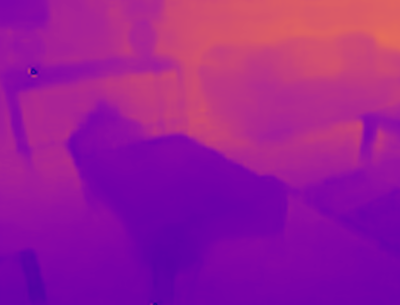}}\\[-0.15in]
			\subfloat{\includegraphics[width = 0.19\linewidth]{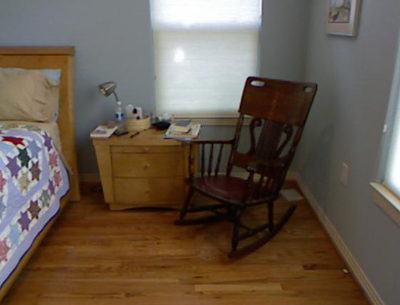}} &
			\subfloat{\includegraphics[width = 0.19\linewidth]{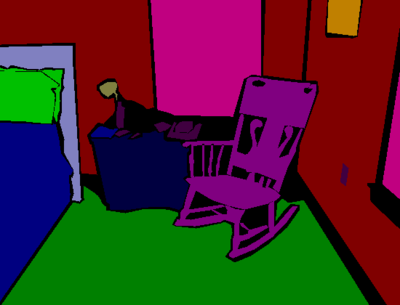}} &
			\subfloat{\includegraphics[width = 0.19\linewidth]{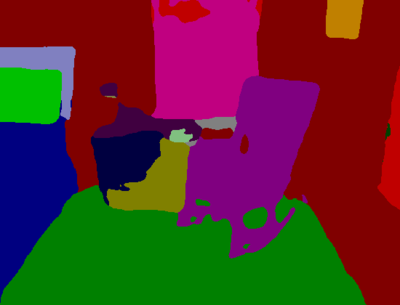}} &
			\subfloat{\includegraphics[width = 0.19\linewidth]{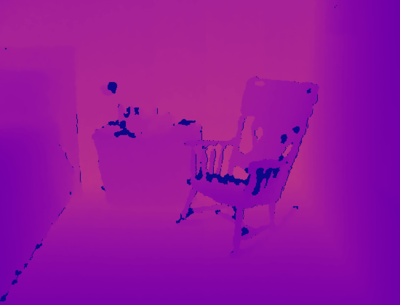}} &
			\subfloat{\includegraphics[width = 0.19\linewidth]{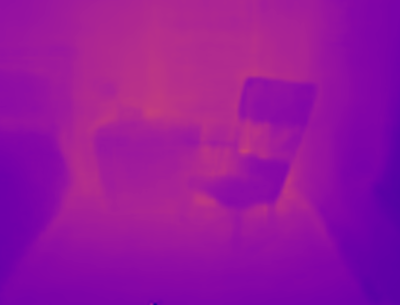}}\\[-0.15in]
			\subfloat{\includegraphics[width = 0.19\linewidth]{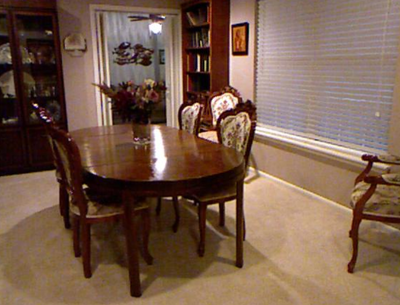}} &
			\subfloat{\includegraphics[width = 0.19\linewidth]{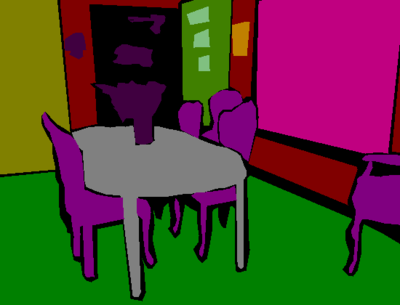}} &
			\subfloat{\includegraphics[width = 0.19\linewidth]{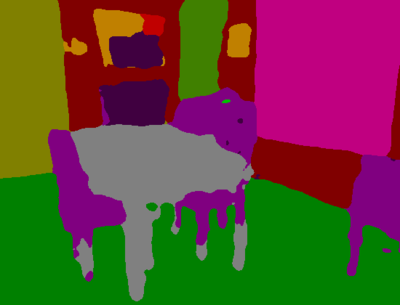}} &
			\subfloat{\includegraphics[width = 0.19\linewidth]{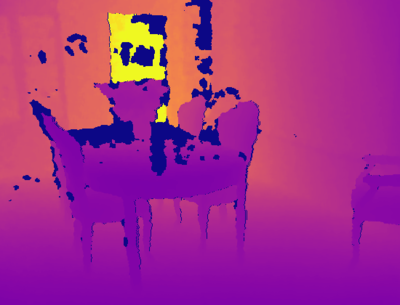}} &
			\subfloat{\includegraphics[width = 0.19\linewidth]{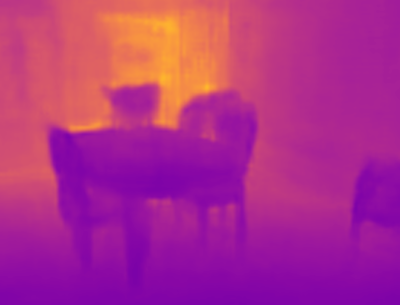}}\\
			Image&GT-Segm&Pred-Segm&GT-Depth&Pred-Depth
		\end{tabular}
	}
	\caption{Qualitative results on the test set of NYUD-v2. The black and dark-blue pixels in `GT-Segm' and `GT-Depth' respectively, indicate pixels without an annotation or label\label{fig:nyud}}
\end{figure}

Quantitatively, we are able to achieve $42.02\%$ mean iou and $0.565$m RMSE (lin) on the validation set~(Table~\ref{table:nyud1}), outperforming several large models, while performing both tasks in real-time simultaneously. More detailed results for depth estimation are given in Table~\ref{table:nyud1_ext}, and qualitative results are provided in Fig.~\ref{fig:nyud}.

\setlength{\tabcolsep}{6pt}
\begin{table}[htb]
	\begin{center}
	\caption{Detailed results on the test set of NYUDv2 for the depth estimation task. For the reported RMSE, abs rel and sqr rel the lower the better, whereas for accuracies ($\delta$) the higher the better\label{table:nyud1_ext}}
		\resizebox{0.49\textwidth}{!}{
			\begin{tabular}{l|c|c|c|c}
				\specialrule{.15em}{0em}{0em}
				 & \textbf{Ours} & Laina~\etal~\cite{Laina2016} & Kendall and Gal~\cite{KendallGal2017Uncertainties} & Qi~\etal~\cite{qi2018geonet}\T\B\\
				\specialrule{.1em}{0em}{0em}
				
				RMSE (lin) & 0.565 & 0.573 & \textbf{0.506} & 0.569\T\B\\
				RMSE (log) & 0.205 & \textbf{0.195} & $-$\T\B\\
				abs rel & 0.149 & 0.127 & \textbf{0.11} & $-$\T\B\\
				sqr rel & \textbf{0.105} & $-$ & $-$ & 0.128\T\B\\
				$\delta<1.25$ & 0.790 & 0.811 & 0.817 & \textbf{0.834}\T\B\\
				$\delta<1.25^{2}$ & 0.955 & 0.953 & 0.959 & \textbf{0.960}\T\B\\
				$\delta<1.25^{3}$ & \textbf{0.990} & 0.988 & 0.989 & \textbf{0.990}\B \\
				\specialrule{.15em}{0em}{0em}
			\end{tabular}
		}

	\end{center}
	\vskip -0.35in
\end{table}
\setlength{\tabcolsep}{1.4pt}

\textbf{Ablation Studies.} To evaluate the importance of pre-training using the synthetic annotations and benefits of performing two tasks jointly, we conduct a series of ablation experiments. In particular, we compare three baseline models trained on the small set of $795$ images and three other approaches that make use of additional data - ours with noisy estimates from a larger model, and two methods, one by Kokkinos~\cite{Kokkinos17}, where the gradients are being accumulated until a certain number of examples is seen, and one by Dvornik~\etal~\cite{DvornikSMS17}, where the task branch is updated every time at least one example is seen.

The results of our experiments are given in Table~\ref{table:nyud2}. The first observation that we make is that performing two tasks jointly on the small set does not provide any significant benefits for each separate task, and even substantially harms semantic segmentation. In contrast, having a large set of depth annotations results in valuable improvements in depth estimation and even semantic segmentation, when it is coupled with a clever strategy of accumulating gradients. Nevertheless, none of the methods can achieve competitive results on semantic segmentation, whereas our proposed approach reaches better performance without any changes to the underlying optimisation algorithm.

\setlength{\tabcolsep}{4pt}
\begin{table}[htb]
	\begin{center}
	\caption{Results of ablation experiments on the test set of NYUDv2. \textit{SD} means how many images have a joint pair of annotations - both segmentation (\textit{S}) and depth (\textit{D}); \textit{task update frequency} denotes the number of examples of each task to be seen before performing a gradient step on task-specific parameters;  \textit{base update frequency} is the number of examples to be seen (regardless of the task) before performing a gradient step on shared parameters\label{table:nyud2}}
		\resizebox{0.49\textwidth}{!}{
			\begin{tabular}{l|c|c|Y{3.5em}|Y{3.5em}|c|c}
				\specialrule{.15em}{0em}{0em} 
				&\multicolumn{2}{c|}{\textbf{Annotations}} & \multicolumn{2}{c|}{\textbf{Update Frequency}} & \textbf{Segm.} &\textbf{Depth}\T\B\\
				\specialrule{.1em}{0em}{0em}
				Method & Pre-Training & Fine-Tuning & Task& Base & mIoU,\% & RMSE (lin),\MakeLowercase{m}\T\B\\
				\specialrule{.1em}{0em}{0em}
				Baseline (SD) & $795$SD & $-$ & $1$ & $1$ & $32.48$ & $0.6328$\T\\
				Baseline (S) & $795$S & $-$ & $1$ & $1$ & $34.44$ & $-$\\
				Baseline (D) & $795$D & $-$ & $1$ & $1$ & $-$ & $0.6380$\B\\
				\hline
				BlitzNet~\cite{DvornikSMS17} & $25405$D + $795$SD & $795$SD & $1$ & $1$ & $34.82$ & $0.5823$\T\\
				UberNet~\cite{Kokkinos17} & $25405$D + $795$SD & $795$SD & $10$ & $30$ & $35.88$ & $0.5728$\\
				\textbf{Ours} & $25405$SD & $795$SD & $1$ & $1$ & \textbf{42.02} & \textbf{0.5648}\B\\
				\specialrule{.15em}{0em}{0em}
			\end{tabular}
		}

	\end{center}
	\vskip -0.2in
\end{table}
\setlength{\tabcolsep}{1.4pt}

\setlength{\tabcolsep}{4pt}
\begin{table*}[t]
	\begin{center}
	\vskip 0.05in
	\caption{Results on the test set of KITTI-6 for segmentation and KITTI for depth estimation%
			\label{table:kitti}}
			\begin{tabular}{l|c|c|c|c|c|c|c|c}
				\specialrule{.15em}{0em}{0em} 
				&&\textbf{Sem. Segm.} & \multicolumn{2}{c|}{\textbf{Depth Estimation}} & \multicolumn{4}{|c}{\textbf{General}}\T\B\\
				\specialrule{.1em}{0em}{0em}
				Model & Regime & mIoU,\% & RMSE (lin),m & RMSE (log) & Parameters,M & Input Size & GFLOPs & speed,ms (mean/std)\T\B\\
				\specialrule{.1em}{0em}{0em} 
				\textbf{Ours} & Segm,Depth & \textbf{87.02} & \textbf{3.453} & $0.182$ & \textbf{2.99} & 1200x350 & \textbf{6.45} & \textbf{16.9$\pm$0.1} \T\B \\
				\hline
				Fast Res.Forests~\cite{ZuoD17} & Segm & $84.9$ & $-$ & $-$ & $-$ & 1200x350 & $-$ & $106.35$\T\\
				Wang et al.~\cite{WangFU15} & Segm & $74.8$ & $-$ & $-$ & $-$ & $-$ & $-$ & $-$\\
				Garg \cite{garg2016unsupervised} & Depth & $-$ & 5.104 & 0.273 & $-$ & $-$ & $-$ & $-$\\
				Goddard \cite{godard2017unsupervised} & Depth & $-$ & 4.471 & 0.232 & $31$ & 512x256 & $-$ & $35.0$ \\
				Kuznietsov \cite{kuznietsov2017semi} & Depth &$-$ &3.518 & \textbf{0.179} & $-$ & 621x187 & $-$ & $48.0$ \B\\ 
				\specialrule{.15em}{0em}{0em} 
			\end{tabular}

	\end{center}
	\vskip -0.2in
\end{table*}
\setlength{\tabcolsep}{1.4pt}

\subsection{KITTI}

KITTI is an outdoor dataset that contains $100$ images semantically annotated for training (with $11$ semantic classes) and $46$ images for testing~\cite{ros:2015} without ground truth depth maps. Following previous work by~\cite{WangFU15}, we keep only $6$ well-represented classes.

Besides segmentation, we follow~\cite{EigenNIPS} and employ $20000$ images with depth annotations available for training~\cite{GeigerLSU13}, and $697$ images for testing. Due to similarities with the CityScapes dataset~\cite{CordtsORREBFRS16}, we consider ResNet-$38$~\cite{WuSH16e} trained on CityScapes as our teacher network to annotate the training images that have depth but not semantic segmentation. In turn, to annotate missing depth annotations on $100$ images with semantic labels from KITTI-6, we first trained a separate copy of our network on the depth task only, and then used it as a teacher. Note that we abandoned this copy of the network and did not make any further use of it.

\begin{figure}[htb]
	\centering
	\begin{tabular}{ccc}
		\subfloat{\includegraphics[width = 0.33\linewidth]{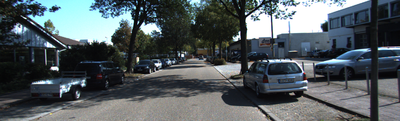}} &
		\subfloat{\includegraphics[width = 0.33\linewidth]{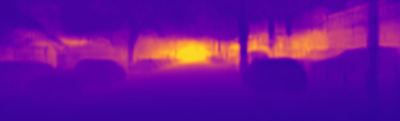}} &
		\subfloat{\includegraphics[width = 0.33\linewidth]{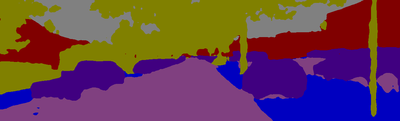}}\\[-0.15in]
		\subfloat{\includegraphics[width = 0.33\linewidth]{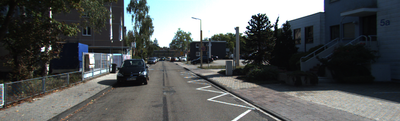}} &
		\subfloat{\includegraphics[width = 0.33\linewidth]{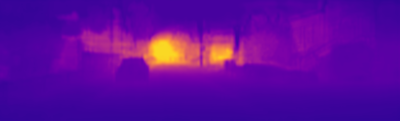}} &
		\subfloat{\includegraphics[width = 0.33\linewidth]{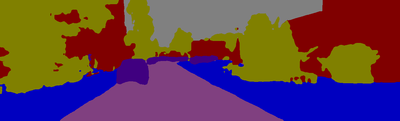}}\\[-0.15in]
		\subfloat{\includegraphics[width = 0.33\linewidth]{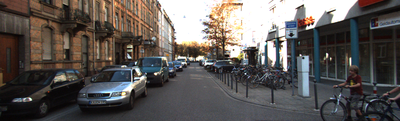}} &
		\subfloat{\includegraphics[width = 0.33\linewidth]{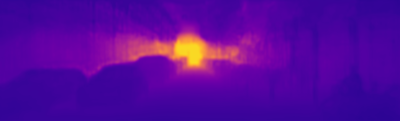}} &
		\subfloat{\includegraphics[width = 0.33\linewidth]{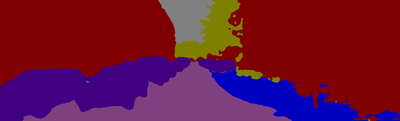}}\\
		Image&Pred-Depth&Pred-Segm
	\end{tabular}
	\caption{Qualitative results on the test set of KITTI (for which only GT depth maps are available). We do not visualise GT depth maps due to their sparsity\label{fig:kitti-depth}}
	\vskip -0.15in
\end{figure}

After pre-training on the large set, we fine-tune the model on the small set of $100$ examples. Our quantitative results are provided in Table~\ref{table:kitti}, while visual results can be seen on Fig.~\ref{fig:kitti-depth}. Per-class segmentation results are given in Table~\ref{table:kitti_ext}. As evident, our approach outperforms other competing methods across a large set of metrics - both on semantic segmentation and depth estimation, while being light-weight and having low latency.

\setlength{\tabcolsep}{4pt}
\begin{table}[htb]
	\begin{center}
	\caption{Detailed segmentation results on the test set of KITTI-6\label{table:kitti_ext}}
		\resizebox{0.49\textwidth}{!}{
			\begin{tabular}{l|c|c|c|c|c|c|c}
				\specialrule{.15em}{0em}{0em}
				Model & sky & building & road & sidewalk & vegetation & car & Total\T\B\\
				\specialrule{.1em}{0em}{0em} 
				\textbf{Ours} & $85.1$ & \textbf{87.7} & \textbf{92.8} & \textbf{82.7} & $86.1$ & \textbf{87.6} & \textbf{87.0}\T\B\\
				\hline 
				Fast Res.Forests~\cite{ZuoD17} & $84.5$ & $85.9$ & $92.3$ & $78.8$ & \textbf{87.8} & $80.3$ & $84.9$\T\\
				Wang et al.~\cite{WangFU15} & \textbf{88.6} & $80.1$ & $80.9$ & $43.6$ & $81.6$ & $63.5$ & $74.8$\B\\
				\specialrule{.15em}{0em}{0em} 
			\end{tabular}
		}

	\end{center}
	\vskip -0.25in
\end{table}
\setlength{\tabcolsep}{1.4pt}

\section{Extensions}
\label{sec:extensions}

The goal of this section is to demonstrate the ease with which our approach can be directly applied in other practical scenarios, such as, for example, the deployment of a single model performing three tasks at once, and the deployment of a single model performing two tasks at once under two different scenarios - indoor and outdoor. As the third task, here we consider surface normals estimation, and as two scenarios, we consider training a single model on both NYUD and KITTI simultaneously without the necessity of having a separate copy of the same architecture for each dataset.

In this section, we strive for simplicity and do not aim to achieve high performance numbers, thus we directly apply the same training scheme as outlined in the previous section.

\subsection{Single Model - Three Tasks}
\label{subsec:ext1}

\begin{figure*}
	\centering
	\resizebox{\textwidth}{!}{
	\begin{tabular}{ccccccc}
		\subfloat{\includegraphics[width = 0.14\linewidth]{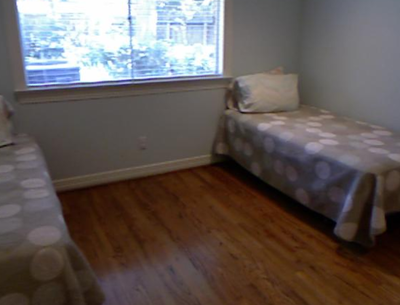}} &
		\subfloat{\includegraphics[width = 0.14\linewidth]{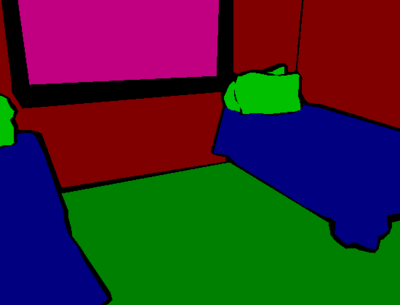}} &
		\subfloat{\includegraphics[width = 0.14\linewidth]{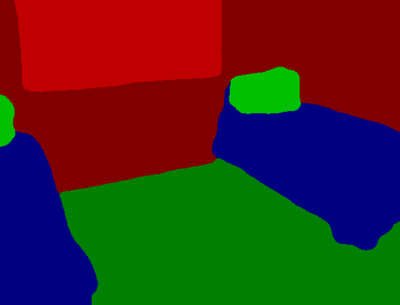}} &
		\subfloat{\includegraphics[width = 0.14\linewidth]{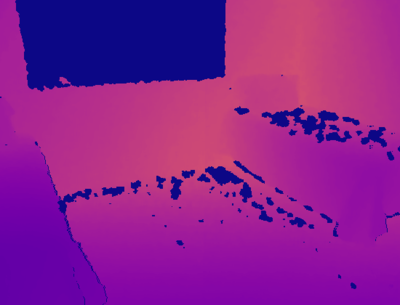}} &
		\subfloat{\includegraphics[width = 0.14\linewidth]{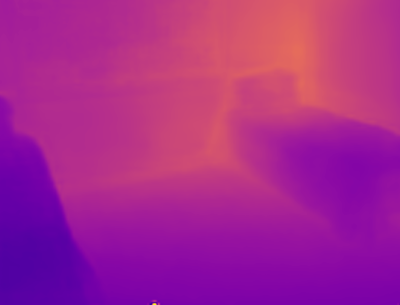}}&
		\subfloat{\includegraphics[width = 0.14\linewidth]{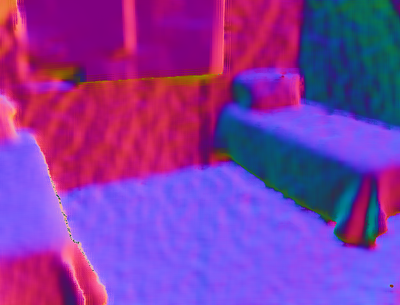}} &
		\subfloat{\includegraphics[width = 0.14\linewidth]{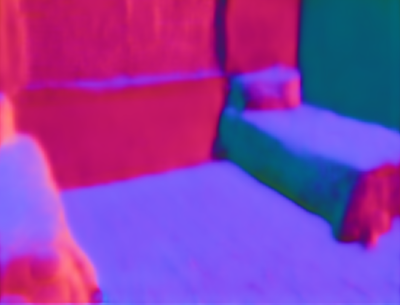}}\\[-0.15in]
		\subfloat{\includegraphics[width = 0.14\linewidth]{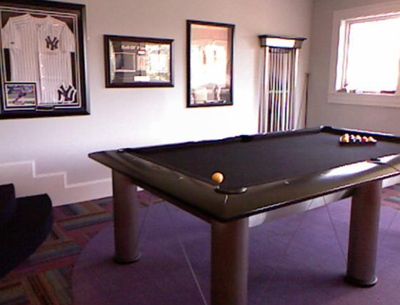}} &
		\subfloat{\includegraphics[width = 0.14\linewidth]{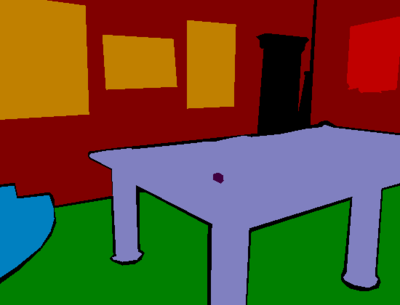}} &
		\subfloat{\includegraphics[width = 0.14\linewidth]{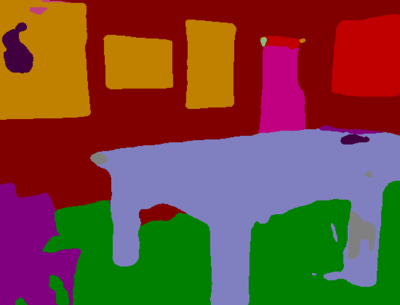}} &
		\subfloat{\includegraphics[width = 0.14\linewidth]{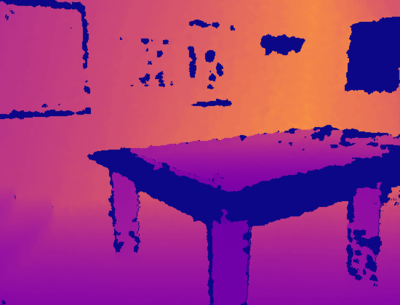}} &
		\subfloat{\includegraphics[width = 0.14\linewidth]{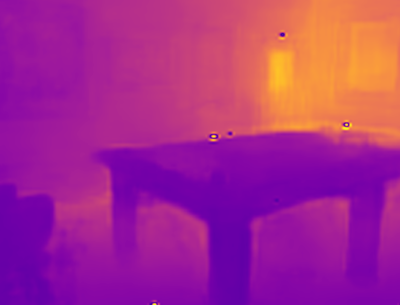}}&
		\subfloat{\includegraphics[width = 0.14\linewidth]{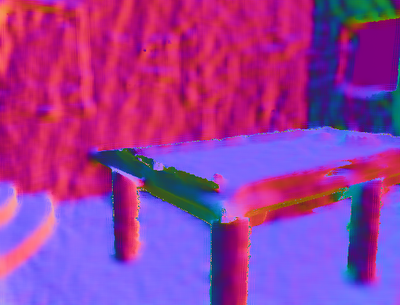}} &
		\subfloat{\includegraphics[width = 0.14\linewidth]{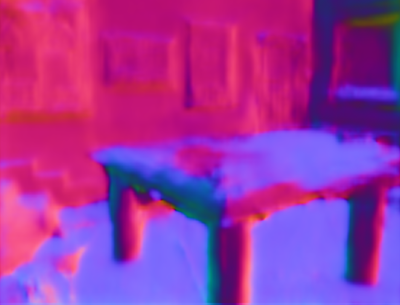}}\\
		Image&GT-Segm&Pred-Segm&GT-Depth&Pred-Depth&GT-Normals&Pred-Normals
	\end{tabular}
	}
	\caption{Qualitative results on the test set of NYUD-v2 for three tasks. The black pixels in the `GT-Segm' images indicate those without a semantic label, whereas the dark blue pixels in the `GT-Depth' images indicate missing depth values\label{fig:nyud3}}
	\vskip -0.2in
\end{figure*}

Analogously to the depth and segmentation branches, we append the same structure with two convolutional layers for surface normals. We employ the negative dot product (after normalisation) as the training loss for surface normals, and we multiply the learning rate for the normals parameters by $10$, as done in~\cite{EigenF15}.

We exploit the raw training set of NYUDv2~\cite{SilbermanHKF12} with more than $300,000$ images, having (noisy) depth maps from the Kinect sensor and with surface normals computed using the toolbox provided by the authors. To acquire missing segmentation labels, we repeat the same procedure outlined in the main experiments - in particular, we use the Light-Weight RefineNet-152 network~\cite{nekrasovlight} to get noisy labels. After pre-training on this large dataset, we divide the learning rate by $10$ and fine-tune the model on the small dataset of $795$ images having annotations for each modality. For surface normals, we employ the annotations provided by Silberman~\etal~\cite{SilbermanHKF12}.

Our straightforward approach achieves practically the same numbers on depth estimation, but suffers a significant performance drop on semantic segmentation (Table~\ref{table:nyud3t}). This might be directly caused by the excessive number of imperfect and noisy labels, on which the semantic segmentation part is being pre-trained. Nevertheless, the results on all three tasks remain competitive, and we are able to perform all three of them in real-time simultaneously. We provide a few examples of our approach on Figure~\ref{fig:nyud3}.

\setlength{\tabcolsep}{4pt}
\begin{table}[htb]
	\begin{center}
	\caption{Results on the test set of NYUDv2 of our single network predicting three modalities at once with surface normals annotations from~\cite{SilbermanHKF12}. The speed of a single forward pass is measured on $640\times480$ inputs. Baseline results (with a single network performing only segmentation and depth) are in \textbf{bold} \label{table:nyud3t}}
		\resizebox{0.49\textwidth}{!}{
			\begin{tabular}{c|c|c|c|c|c}
				\specialrule{.15em}{0em}{0em}
				\textbf{Segm.} & \multicolumn{2}{c|}{\textbf{Depth}} &  \multicolumn{2}{c|}{\textbf{Surface Normals}} & \multicolumn{1}{c}{\textbf{General}}\T\B\\
				\specialrule{.1em}{0em}{0em}
				mIoU,\% & RMSE (lin),m & RMSE (log) & Mean Angle & Median Angle & speed,ms (mean/std)\T\B\\
				\specialrule{.1em}{0em}{0em}
				$38.66$ & $0.566$ & $0.209$ & $23.95$ & $17.74$ & $13.4$$\pm$$0.1$\B\T\\
				\specialrule{.1em}{0em}{0em}
				\textbf{42.02} & \textbf{0.565} & \textbf{0.205} & $-$ & $-$ & \textbf{12.8}$\pm$\textbf{0.1}\T\B\\
				\specialrule{.15em}{0em}{0em}
			\end{tabular}
		}

	\end{center}
	\vskip -0.25in
\end{table}
\setlength{\tabcolsep}{1.4pt}

\subsection{Single Model - Two Datasets, Two Tasks}
\label{subsec:ext2}

Next, we consider the case when it is undesirable to have a separate copy of the same model architecture for each dataset. Concretely, our goal is to train a single model that is able to perform semantic segmentation and depth estimation on both NYUD and KITTI at once. To this end, we simply concatenate both datasets and amend the segmentation branch to predict $46$ labels ($40$ from NYUD and $6$ from KITTI-6).

We follow the exact same training strategy, and after pre-training on the union of large sets, we fine-tune the model on the union of small training sets.
Our network exhibits no difficulties in differentiating between two regimes (Table~\ref{table:nyud-kitti}), and achieves results at the same level with the separate approach on each of the datasets without a substantial increase in model capacity.

\setlength{\tabcolsep}{6pt}
\begin{table}[htb]
	\begin{center}
	\caption{Results on the test set of NYUDv2, KITTI (for depth) and KITTI-6 (for segmentation) of our single network predicting two modalities on both datasets together. Baseline results (with separate networks per dataset) are in \textbf{bold}\label{table:nyud-kitti}}
		\resizebox{0.49\textwidth}{!}{
			\begin{tabular}{c|c|c|c|c|c}
				\specialrule{.15em}{0em}{0em}
				\multicolumn{3}{c|}{\textbf{NYUDv2}} \T\B& %
				\multicolumn{3}{|c}{\textbf{KITTI}}\\
				\specialrule{.1em}{0em}{0em}
				\textbf{Segm.} & \multicolumn{2}{c|}{\textbf{Depth}} & \textbf{Segm.} &  \multicolumn{2}{|c}{\textbf{Depth}}\T\B\\
				\specialrule{.1em}{0em}{0em}
				mIoU,\% & RMSE (lin),m &  RMSE (log) & mIoU,\% & RMSE (lin),m &  RMSE (log)\T\B\\
				\specialrule{.1em}{0em}{0em}
				$38.76$ & $0.59$ & $0.213$ & $86.1$ & $3.659$ & $0.190$\T\B\\
				\specialrule{.1em}{0em}{0em}
				\textbf{42.02} & \textbf{0.565} & \textbf{0.205} & \textbf{87.0} & \textbf{3.453} & \textbf{0.182}\T\B\\
				\specialrule{.15em}{0em}{0em}
			\end{tabular}
		}

	\end{center}
	\vskip -0.25in
\end{table}
\setlength{\tabcolsep}{1.4pt}

\subsection{Dense Semantic SLAM}
\label{subsec:ext3}
Finally, we demonstrate that quantities predicted by our joint network performing depth estimation and semantic segmentation indoors can be directly incorporated into existing SLAM frameworks.

In particular, we consider SemanticFusion~\cite{McCormacHDL17}, where the SLAM reconstruction is carried out by ElasticFusion~\cite{whelan2016elasticfusion}, which relies on RGB-D inputs in order to find dense correspondences between frames. A separate CNN, also operating on RGB-D inputs, was used by McCormac~\etal~\cite{McCormacHDL17} to acquire $2$D semantic segmentation map of the current frame. A dense $3$D semantic map of the scene is obtained with the help of tracked poses predicted by the SLAM system.

\begin{figure}
	\centering
	\begin{tabular}{cc}
		\includegraphics[width=0.19\textwidth]{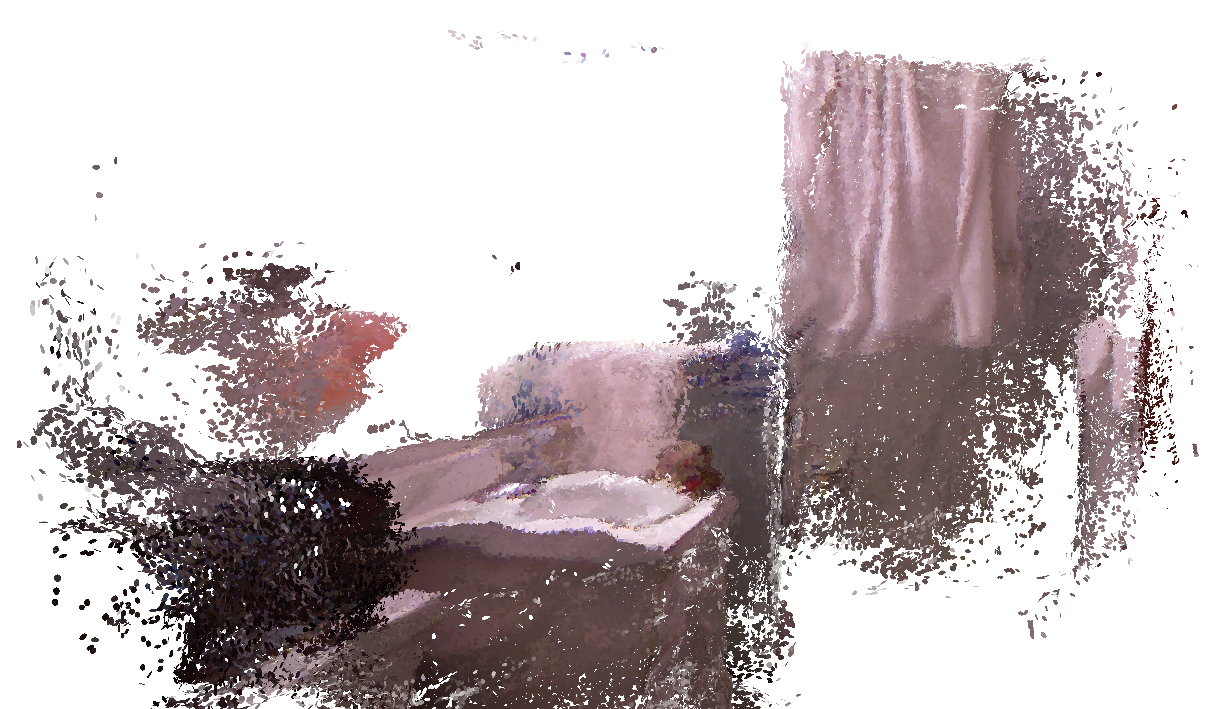} &
		\includegraphics[width=0.19\textwidth]{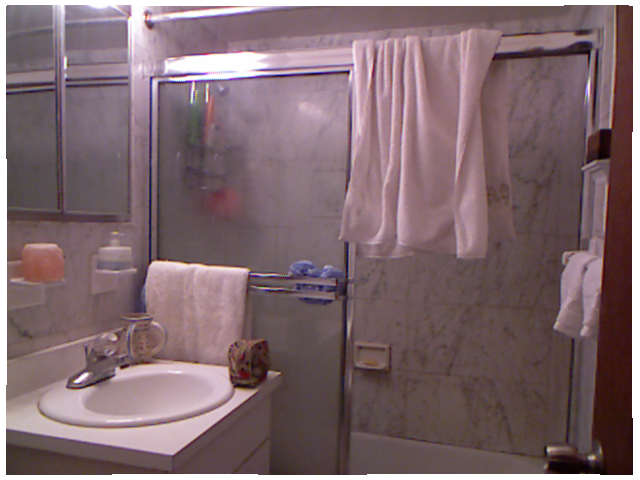}\\
		Point Cloud (ours) & RGB Frame \\
		\includegraphics[width=0.19\textwidth]{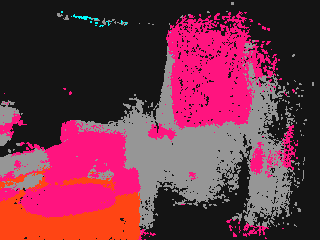} &
		\includegraphics[width=0.19\textwidth]{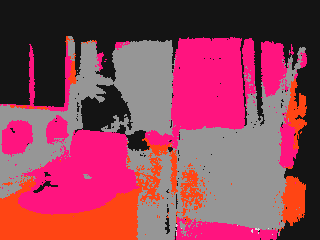}\\
		Segm. Map (ours) & Segm. Map~\cite{McCormacHDL17}
	\end{tabular}
	\caption{3D reconstruction output using our per-frame depths and segmentation inside SemanticFusion~\cite{McCormacHDL17}
	\label{fig:semfus}}
	\vskip -0.2in
\end{figure}

We consider one sequence of the NYUD validation set provided by the authors\footnote{\url{https://bitbucket.org/dysonroboticslab/semanticfusion/overview}}, and directly replace ground truth depth measurements with the outputs of our network performing depth and segmentation jointly (Sect.~\ref{exp-nyudv2}). Likewise, we do not make use of the authors' segmentation CNN and instead exploit segmentation predictions from our network. Note also that our segmentation network was trained on $40$ semantic classes, whereas here we directly re-map the results into the $13$-classes domain~\cite{couprie2013indoor}. We visualise dense surfel-based reconstruction along with dense segmentation and current frame on Fig.~\ref{fig:semfus}. Please refer to the supplementary video material\footnote{\url{https://youtu.be/qwShIBhaq8Y}} for the full sequence results.

\section{Conclusion}
\label{sec:conclusion}

We believe that efficient and effective exploitation of visual information in robotic applications using deep learning models is crucial for further development and deployment of robots and autonomous vehicles. To this end, we presented a simple way of achieving real-time performance for the joint task of depth estimation and semantic segmentation. We showcased that it is possible (and indeed beneficial) to re-use large existing models in order to generate synthetic labels important for the pre-training stage of a compact model. Moreover, our method can be easily extended to handle more tasks and more datasets simultaneously, while raw depth and segmentation predictions of our network can be seamlessly used within available dense SLAM systems. As our future work, we will consider whether it would be possible to directly incorporate expert's uncertainty during the pre-training stage to acquire better results, as well as the case when there is no reliable expert available. Another interesting direction lies in incorporating findings of Zamir~\etal~\cite{zamir2018taskonomy} in order to reduce the total number of training annotations without sacrificing performance.

\section*{Acknowledgements}
The authors would like to thank the anonymous reviewers for their helpful and constructive comments. This research was supported by the Australian Research Council through the Australian Centre for Robotic Vision (CE140100016), the ARC Laureate Fellowship FL130100102 to IR, and the HPC cluster Phoenix at the University of Adelaide.

\bibliographystyle{IEEEtran}
\bibliography{example.bib}

\end{document}